\documentclass{bmvc2k}

\newcommand{\bg}[1]{\boldsymbol{#1}} %Bold Greek letters
\newcommand{\bm}[1]{\mathbf{#1}} %Bold vectors and matrices

\newcommand\raiseT[2]{%
\setbox0\hbox{$#1{#2}$}\raise\dp0\box0}

\usepackage{amsmath,amssymb}
\usepackage{booktabs}
\usepackage{wrapfig}

\title{PEEKABOO: Hiding Parts of an Image for Unsupervised Object Localization}

\addauthor{Hasib Zunair}{hasibzunair@gmail.com}{1}
\addauthor{A. Ben Hamza}{hamza@ciise.concordia.ca}{1}
\addinstitution{
 Concordia University \\
 Montreal, QC, Canada
}

\runninghead{Zunair, Hamza}{PEEKABOO: Unsupervised Object Localization}

\def\etal{\emph{et al}\bmvaOneDot}

\begin{document}

\maketitle

\begin{abstract}
Localizing objects in an unsupervised manner poses significant challenges due to the absence of key visual information such as the appearance, type and number of objects, as well as the lack of labeled object classes typically available in supervised settings. While recent approaches to unsupervised object localization have demonstrated significant progress by leveraging self-supervised visual representations, they often require computationally intensive training processes, resulting in high resource demands in terms of computation, learnable parameters, and data. They also lack explicit modeling of visual context, potentially limiting their accuracy in object localization. To tackle these challenges, we propose a single-stage learning framework, dubbed PEEKABOO, for unsupervised object localization by learning context-based representations at both the pixel- and shape-level of the localized objects through image masking. The key idea is to selectively hide parts of an image and leverage the remaining image information to infer the location of objects without explicit supervision. The experimental results, both quantitative and qualitative, across various benchmark datasets, demonstrate the simplicity, effectiveness and competitive performance of our approach compared to state-of-the-art methods in both single object discovery and unsupervised salient object detection tasks. Code and pre-trained models are available at: \textcolor{blue}{https://github.com/hasibzunair/peekaboo}
\end{abstract}

%Keywords:

\section{Introduction}
Localizing objects in images or videos is a fundamental task in many real-world computer vision systems, including autonomous driving and robotics. The goal is to detect/segment the most salient objects in an image. High-performing methods for object localization typically rely on large cohorts of human-annotated datasets for supervised learning~\cite{lin2014microsoft}. However, this learning paradigm faces two main significant limitations. First, acquiring annotated datasets is notoriously time-consuming and costly, as well as prone to errors due in part to annotators' fatigue. Second, the finite and predefined nature of object classes limits the scope of what supervised models can localize, rendering them ineffective for segmenting novel objects.

In recent years, several approaches have been proposed to tackle the daunting task of unsupervised object localization, which is inherently challenging. This difficulty arises from the need to identify salient objects without prior knowledge of their appearance or reliance on human-annotated datasets. Some approaches for unsupervised object discovery leverage region proposal techniques and representer point selection~\cite{uijlings2013selective,zitnick2014edge,Song2023point}, while others rely on adversarial training strategies~\cite{melas2021finding,voynov2021object}. Moreover, some methods refine mask proposals~\cite{wang2022freesolo,zadaianchuk2022unsupervised}, albeit at the expense of requiring millions of learnable parameters, extensive unlabeled image datasets, and multi-stage training procedures. Drawing upon the localization properties of self-supervised visual representations~\cite{caron2021emerging}, training-free graph-based methods~\cite{simeoni2021localizing,wang2022self,melas2022deep} have demonstrated robust performance across various tasks and benchmarks. However, these approaches often struggle in practical real-world scenarios due largely to variations in illumination, reflections on water and shiny surfaces~\cite{simeoni2023unsupervisedd}, overlapping objects~\cite{simeoni2021localizing}, occlusions of small objects, complex backgrounds, assumptions that the largest object is salient, and segmenting multiple objects as one~\cite{wang2022self,melas2022deep,lv2023weakly}. Furthermore, they may erroneously count multiple instances of the same object as one, leading to over- or under-segmentation, as well as segment non-salient regions and produce noisy and discontinuous predictions. In addition, training-based methods impose significant computational demands~\cite{shin2022unsupervised}, rely on extensive unlabeled image datasets~\cite{seitzer2022bridging}, employ combinations of multiple networks, ensemble methods, and require test-time training~\cite{shin2022unsupervised,aflalo2023deepcut}. Overcoming these challenges is essential for deploying reliable vision systems in real-world scenarios, such as autonomous driving, where accuracy and efficiency are paramount. Therefore, there is a pressing need to develop unsupervised object localization methods that are both accurate and computationally efficient.

To address the aforementioned challenges, we introduce PEEKABOO\footnote{The name PEEKABOO is inspired by the children's game of hiding one's face and then revealing it suddenly.}, a novel single-stage learning framework for unsupervised object localization by adopting a context-based representation learning strategy to enhance foreground-background discrimination, without resorting to pixel-level reconstruction. Given an image and its heavily masked counterpart, our model first predicts pixel-level semantic labels for both the unmasked and masked inputs. This process exploits contextual information by utilizing nearby non-masked pixels to inform predictions for masked pixels, followed by learning shape-level context by ensuring consistency between the object mask predictions for the masked and unmasked inputs. The main contributions of this work can be summarized as follows: (1) We propose a learning paradigm that aims to explicitly model context through hiding parts of an image for unsupervised object localization; (2) We show through experimental results and ablation studies on six benchmark datasets that PEEKABOO yields competitive performance against state-of-the-art methods in both single object discovery and unsupervised saliency detection tasks.
%\item We demonstrate that PEEKABOO is, by orders of magnitude, efficient in terms of computational power, model size and data.

\section{Related work}
\noindent\textbf{Unsupervised Object Localization.}\quad Generative representation learning methods like the one proposed by Melas-Kyriazi \etal~\cite{melas2021finding} employ a two-stage adversarial training approach, utilizing BigBiGAN~\cite{donahue2019large} to create a synthetic dataset, followed by segmenter training. Another line of work  involves generating mask proposals that are refined through training. FreeSOLO~\cite{wang2022freesolo} proposes Free Mask to generate coarse object masks refined via self-training, but requires several millions of learnable parameters and large-scale unlabeled image datasets. More recent efforts leverage the strong localization properties of DINO~\cite{caron2021emerging}, a self-supervised pre-training method, with or without additional training. For instance, LOST~\cite{simeoni2021localizing} is training-free and constructs a weighted graph of patches, segmenting foreground objects based on their similarity, but struggles in scenarios involving object overlap or when objects cover most of the image. TokenCut~\cite{wang2022self} and Deep Spectral Methods (DSMs)~\cite{melas2022deep} are graph-based methods, but their performance is hindered by the computational overhead of computing the eigenvectors of the Laplacian matrix. They often assume a single salient object occupies the foreground and may count multiple instances of the same object as one. Moreover, they struggle with small and occluded objects, leading to over- or under-segmentation. SelfMask~\cite{shin2022unsupervised} is training-based method, which employs an ensemble of three self-supervised vision encoders (SwAV~\cite{caron2020unsupervised}, DINO~\cite{caron2021emerging}, MoCov2~\cite{chen2020improved}) and spectral clustering to generate more than twenty salient masks for each image, making the training considerably more expensive. DINOSAUR~\cite{seitzer2022bridging} learns representations from hundreds of thousands of unlabeled images by separating the features of an image and reconstructing them into individual objects or parts. WSCUOD~\cite{lv2023weakly} proposes a semantic-guided method that extracts object-centric representations to generate the final object regions from the image. DeepCut~\cite{aflalo2023deepcut} utilizes features extracted by DINO~\cite{caron2021emerging} and trains a graph neural network to generate segmentation masks, incorporating test-time training on each image. In contrast to these approaches, our method avoids heavy and complex training procedures such as large-scale adversarial training, multi-stage training, learning millions of parameters, and test-time training.

\smallskip\noindent\textbf{Masked Image Modeling for Vision.}\quad While early approaches for masked image modeling focused on reconstructing missing parts from masked images~\cite{pathak2016context, liu2018image}, recent methods such as Masked Autoencoders~\cite{he2021masked} and DINO~\cite{caron2021emerging} have aimed to predict visual tokens in a self-supervised manner~\cite{bao2021beit}. Also, masking has recently been utilized as supervisory signals for models to learn context-based representations in fully supervised learning settings, facilitating accurate perception of 2D~\cite{zunair2024learning} and 3D~\cite{liu2023cpcm} visual data. However, the application of masked modeling tailored for unsupervised learning, specifically for discovering novel objects from unlabeled data, remains relatively unexplored. Moreover, our approach distinguishes itself from existing methods by making pixel-level predictions instead of predicting missing parts of the input itself (i.e., pixel or visual token).

\section{Method}
\begin{figure*}[!htp]
\centering
\includegraphics[scale=0.65]{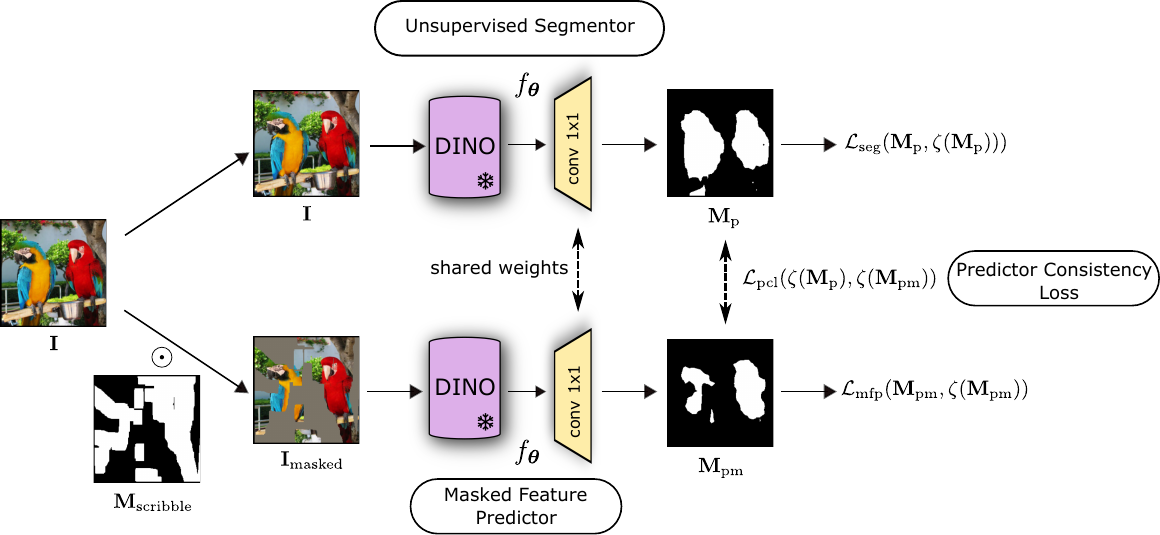}
\caption{\textbf{Overview of PEEKABOO framework for unsupervised object localization}. The proposed learning paradigm consists of an Unsupervised Segmentor, a Masked Feature Predictor and a Predictor Consistency Loss. Here, $f_{\bg{\theta}}$ is a frozen Self-distillation with No Labels (DINO) encoder~\cite{caron2021emerging} paired with a lightweight trainable $1\times 1$ convolutional layer decoder having only 770 learnable parameters. The two branches are identical and share weights. After training, $f_{\bg{\theta}}$ is utilized to generate class-agnostic predicted segmentation masks.}
\label{Fig:schema}
\end{figure*}

We begin by formulating the task at hand and subsequently introduce the main components of the proposed masked unsupervised object localization framework, as illustrated in Figure~\ref{Fig:schema}.

\smallskip\noindent\textbf{Problem Statement.}\quad We address two main tasks in unsupervised object localization from images: single object discovery and unsupervised saliency detection. The former involves enclosing the main object or one of the main objects of interest within a bounding box, while the latter aims to generate a binary mask where foreground objects are assigned a value of 1 and the background is assigned 0, thereby highlighting the object(s) of interest in the image. Both tasks are class-agnostic, i.e., they are not restricted to a predefined set of object classes.

\subsection{Masked Unsupervised Object Localization}
\noindent\textbf{Self-Training Unsupervised Segmenter.} \quad We design an unsupervised segmenter model $f_{\bg{\theta}}$ by leveraging the frozen Self-distillation with No Labels (DINO)~\cite{caron2021emerging} as an encoder for feature extraction with a lightweight single $1\times 1$ convolutional layer decoder having only 770 learnable parameters. The model is trained end-to-end in a self-supervised fashion, and learns to predict a refined version of its own predictions from the input image $\bm{I}$. The predicted mask $\bm{M}_\text{p}$ is refined by applying image binarization, followed by bilateral solver~\cite{barron2016fast}, an edge-aware smoothing technique that enhances mask quality. The refined mask, denoted as $\zeta(\bm{M}_\text{p})$, serves as pseudo ground-truth for self-training, with $\zeta(\cdot)$ representing the image binarization and bilateral solver operation. We train $f_{\bg{\theta}}$ by minimizing the segmenter loss $\mathcal{L}_\text{seg} (\bm{M}_\text{p}, \zeta(\bm{M}_\text{p}))$, which is the binary cross-entropy between the predicted mask and its refined version. Application-specific loss functions can also be used in lieu of cross-entropy to further improve performance depending on the use-case.

\smallskip\noindent\textbf{Masked Input.}\quad Incorporating image masking into context-based representation learning is intuitive as the masked image retains some contextual information, requiring the model to capture background knowledge about the image. To facilitate this, we leverage the Irregular Mask (IMs) dataset~\cite{liu2018image}, which is a collection of masks with random streaks and holes of arbitrary shapes commonly used for image inpainting~\cite{liu2018image,suvorov2022resolution}. From this dataset, we create a subset of images containing only large masks by filtering the masks based on the percentage $p$ of zero pixels they contain. Masks with $p$ greater than 50\% are included, while others are discarded. In our experiments, we observe that these large masks generally enhance localization performance compared to smaller ones. During PEEKABOO training, we randomly sample a large mask, perform binary thresholding to set pixel values to either 0 or 1, and denote this mask as $\bm{M}_\text{scribble}$. We generate a masked image $\bm{I}_\text{masked} = \bm{I}\odot \bm{M}_\text{scribble}$, where $\bm{I}$ is the original input image and $\odot$ denotes element-wise multiplication. Essentially, the masked image retains a layout akin to the original input, albeit with substantial portions of the pixels removed.

\smallskip\noindent\textbf{Masked Feature Predictor (MFP).}\quad  We introduce MFP to explicitly learn context-based representations. MFP is tasked with generating pixel-level semantic labels, distinguishing between foreground and background, for the masked portions of the image. This prompts the model to learn contextual information by utilizing the available pixel-level details of neighboring non-masked pixels to predict the label for each masked pixel. Given the masked image $\bm{I}_\text{masked}$, we train $f_{\bg{\theta}}$ to produce the predicted mask $\bm{M}_\text{pm}$. It is important to note that $f_{\bg{\theta}}$ is a Siamese network-like architecture~\cite{bromley1993signature}, where the branches are identical and share weights, and is trained by minimizing the binary cross-entropy loss $\mathcal{L}_\text{mfp} (\bm{M}_\text{pm}, \zeta(\bm{M}_\text{pm}))$.

\smallskip\noindent\textbf{Predictor Consistency Loss (PCL).}\quad The objective of PCL is to ensure consistency between predictions for the input image and its masked version. By aligning these predictions, PCL learns shape-level representations that complement the pixel-level representations learned by MFP. This alignment is achieved by ensuring that the predictions of the masked input match those of the unmasked input. Since both predictions are generated by the same model $f_{\bg{\theta}}$, this fosters invariance to images with partial inputs, ultimately leading to improved localization performance. More specifically, we maximize the similarity between the predictions from the unsupervised segmenter and masked feature predictor by minimizing the $L_2$-loss $\mathcal{L}_\text{pcl}=\Vert\zeta(\bm{M}_\text{p})- \zeta(\bm{M}_\text{pm})\Vert^{2}$.

\smallskip\noindent\textbf{Overall Loss Function.}
Using the unsupervised segmenter, masked feature predictor and predictor consistency loss, we define the overall loss function as follows:
\begin{equation}
\mathcal{L}_\text{total} = \alpha \mathcal{L}_\text{seg} (\bm{M}_\text{p}, \zeta(\bm{M}_\text{p})) +  \mathcal{L}_\text{mfp} (\bm{M}_\text{pm}, \zeta(\bm{M}_\text{pm})) + \mathcal{L}_\text{pcl} (\zeta(\bm{M}_\text{p}), \zeta(\bm{M}_\text{pm})),
\label{eq:loss_final}
\end{equation}
where the scalar $\alpha$ is a nonnegative trade-off hyperparameter, which controls the contribution of the segmenter loss and is set to 3/2. During training, $\mathcal{L}_\text{total}$ is minimized between predictions and pseudo ground-truth labels using stochastic gradient descent for several epochs to learn the parameters of $f_{\bg{\theta}}$. Similar to~\cite{shin2022unsupervised,simeoni2023unsupervisedd}, we use only $10,553$ images from the DUTS-TR dataset~\cite{wang2017learning} for training PEEKABOO. For inference, given a test image $\bm{I}$, the trained network $f_{\bg{\theta}}$ is employed in unsupervised object localization to obtain segmentation masks.

\section{Experiments}
In this section, we evaluate the performance of PEEKABOO in comparison with state-of-the-art methods. Details on the implementation, architecture and training, as well as method cost discussion and additional experimental results are included in the supplementary material.

\subsection{Experimental Setup}
\noindent\textbf{Datasets.}\quad For single object discovery, we conduct experiments on PASCAL VOC07~\cite{everingham2008pascal} and VOC12~\cite{everingham2012pascal}, which consist images of single large object, and also on COCO20K~\cite{lin2014microsoft}, which includes two and often dozens of objects. For the unsupervised saliency detection task, we evaluate the performance of our approach on three popular saliency detection benchmark datasets: DUT-OMRON~\cite{yang2013saliency}, DUTS-TE~\cite{wang2017learning} and ECSSD~\cite{shi2015hierarchical}. These datasets encompass a diverse array of objects set against various backgrounds.

\smallskip\noindent\textbf{Baselines.}\quad We evaluate the performance of our method against several state-of-the-art
training-free methods, including LOST~\cite{simeoni2021localizing}, TokenCut~\cite{wang2022self} and Deep Spectral Methods~\cite{melas2022deep}. We also compare against training-based generative methods like BigBiGAN~\cite{voynov2021object} and Melas-Kyriazi et al.~\cite{melas2021finding}, along with Transformer-based methods such as FOUND~\cite{simeoni2023unsupervisedd}, SelfMask~\cite{shin2022unsupervised}, FreeSOLO~\cite{wang2022freesolo}, and DINOSAUR~\cite{seitzer2022bridging}. Furthermore, we compare against more recent methods such as WSCUOD~\cite{lv2023weakly} and DeepCut~\cite{aflalo2023deepcut}. Notably, these recent approaches typically entail multiple stages of training, learning several millions of parameters, test-time training, combining multiple learnable networks, model ensembling, and require training on hundreds of thousands of images.

\smallskip\noindent\textbf{Evaluation Metrics.}\quad For single object discovery, we report results using the Correct Localization (CorLoc) metric~\cite{simeoni2021localizing,simeoni2023unsupervisedd}, which measures the percentage of correct boxes, where predicted boxes have an intersection-over-union (IoU) greater than $0.5$ with one of the ground-truth boxes. For unsupervised saliency detection, we report results in terms of IoU, pixel wise accuracy (Acc) and maximal $F_{\beta}$ score (max $F_{\beta}$).

\subsection{Results and Analysis}
\noindent\textbf{Single Object Discovery Results.}\quad We evaluate the performance of PEEKABOO against several state-of-the-art methods on three single object discovery datasets, and the results are summarized in Table~\ref{table:sota_obj_discovery}. We use PEEKABOO with a ViT-S/8~\cite{dosovitskiy2020vit} architecture pre-trained with DINO~\cite{caron2021emerging}. As reported in Table~\ref{table:sota_obj_discovery}, PEEKABOO demonstrates superior performance compared to all training-free methods, including LOST~\cite{simeoni2021localizing}, TokenCut~\cite{wang2022self}, and DSM~\cite{melas2022deep}. Moreover, PEEKABOO maintains computational efficiency by avoiding the computation of eigenvectors, thereby ensuring faster inference~\cite{wang2022self,melas2022deep}. This demonstrates the efficiency of PEEKABOO for real-time object discovery. PEEKABOO also performs better than training-based methods (SelfMask~\cite{shin2022unsupervised}, FreeSOLO~\cite{wang2022freesolo}, FOUND~\cite{simeoni2023unsupervisedd}, WSCUOD~\cite{lv2023weakly} and DeepCut~\cite{aflalo2023deepcut}), while being notably simpler to train. In addition, it outperforms DINOSAUR~\cite{seitzer2022bridging} on VOC07 and VOC12, while DINOSAUR~\cite{seitzer2022bridging} performs better on COCO20K. However, it is worth noting that DINOSAUR~\cite{seitzer2022bridging} incurs an extensive training cost, as it employs a significantly larger ViT-based architecture trained on over three hundred thousand images sourced from both synthetic and real-world datasets. Evidently, PEEKABOO outperforms DINOSAUR~\cite{seitzer2022bridging} on VOC12, achieving a relative improvement of 7.8\% in terms of CorLoc. It also performs better than DeepCut~\cite{aflalo2023deepcut}, which employs multiple learnable
networks (ViTs and GNNs) and involves test-time training. PEEKABOO stands out for its accuracy and efficiency, boasting approximately 42,857 times fewer learnable parameters compared to SelfMask~\cite{shin2022unsupervised}, which relies on an ensemble of three self-supervised vision encoders. Similarly, it is significantly more parameter-efficient than FreeSOLO~\cite{wang2022freesolo}, which employs over two hundred thousand images and 66 million learnable parameters, roughly 24 and 85,714 times higher than our method, respectively. These results underscore the accuracy and efficiency of our method in single object discovery.

\begin{table}[!htb]
\setlength{\tabcolsep}{4pt}
\begin{center}
\caption{\textbf{Performance comparison of our method and state-of-the-art methods in single object discovery using CorLoc as evaluation metric}. The best results are in bold, while the second-best are underlined. Higher values indicate better results. ``+CAD'' denotes an additional second-stage class-agnostic detector trained with unsupervised pseudo-box labels.}
\smallskip
\label{table:sota_obj_discovery}
\begin{tabular}{lclll}
\toprule[1pt]
\smallskip
Method & Learning & VOC07 & VOC12 & COCO20K\\
\midrule[1pt]
\noalign{\smallskip}
Zhang et al.~\cite{zhang2020object} & & 46.2 & 50.5 & 34.8\\
DDT+~\cite{wei2019unsupervised} & & 50.2 & 53.1 & 38.2\\
rOSD~\cite{vo2020toward} & & 54.5 & 55.3 & 48.5\\
LOD~\cite{vo2021large} & & 53.6 & 55.1 & 48.5\\
DINO~\cite{caron2021emerging} & & 45.8 & 46.2 & 42.1\\
LOST~\cite{simeoni2021localizing} (ViT-S/16) & & 61.9 & 64.0 & 50.7\\
LOST + CAD~\cite{simeoni2021localizing} & & 65.7 & 70.4 & 57.5\\
DSM~\cite{melas2022deep} (ViT-S/16) & & 62.7 & 66.4 & 52.2\\
TokenCut~\cite{wang2022self} (ViT-S/16) & & 68.8 & 72.1 & 58.8\\
TokenCut + CAD~\cite{wang2022self} & & 71.4 & 75.3 & 62.6\\
SelfMask~\cite{shin2022unsupervised} & \checkmark & \underline{72.3} & 75.3 & 62.7\\
FOUND$\dagger$~\cite{simeoni2023unsupervisedd} & \checkmark & 71.7 & \underline{75.6} & 61.1\\
FreeSOLO~\cite{wang2022freesolo} & \checkmark & 56.1 & 56.7 & 52.8\\
DeepCut~\cite{aflalo2023deepcut} & \checkmark & 69.8 & 72.2 & 61.6\\
WSCUOD~\cite{lv2023weakly} & \checkmark & 70.6 & 72.1 & 63.5\\
DINOSAUR~\cite{seitzer2022bridging} & \checkmark & - & 70.4 & \textbf{67.2}\\
\textbf{PEEKABOO (ViT-S/8) (Ours)} & \checkmark & \textbf{72.7} & \textbf{75.9} & \underline{64.0}\\
\bottomrule[1pt]
\end{tabular}
\end{center}
\end{table}

\smallskip\noindent\textbf{Unsupervised Saliency Detection Results.}\quad In Table~\ref{table:sota_saliency_detection}, we present the performance comparison across three popular saliency detection benchmarks. PEEKABOO demonstrates superiority over training-based techniques such as BigBiGAN~\cite{voynov2021object} and Melas-Kyriazi et al.\cite{melas2021finding}, which rely on large-scale adversarial training. Notably, PEEKABOO consistently outperforms training-free methods, including LOST\cite{simeoni2021localizing}, TokenCut~\cite{wang2022self}, and DSM~\cite{melas2022deep} across all evaluation metrics. Interestingly, PEEKABOO achieves these results with computational efficiency, even when compared to approaches incorporating bilateral solver post-processing~\cite{barron2016fast}. Moreover, it demonstrates superior performance compared to SelfMask~\cite{shin2022unsupervised}, except for IoU on DUT-OMRON, despite the substantial training cost associated with SelfMask. This cost stems from its ensemble approach, utilizing three self-supervised vision encoders (SwAV~\cite{caron2020unsupervised}, DINO~\cite{caron2021emerging}, MoCov2~\cite{chen2020improved}), and generating twenty masks for each image during training. Furthermore, PEEKABOO outperforms DeepCut~\cite{aflalo2023deepcut}, which employs multiple learnable networks (ViTs and GNNs) and requires test-time training. It also consistently outperforms WSCUOD~\cite{lv2023weakly} across all datasets and evaluation metrics, even when WSCUOD employs bilateral solver post-processing. Notably, WSCUOD utilizes weak supervisory signals for training, whereas PEEKABOO operates directly on unlabeled images and has significantly fewer learnable parameters, approximately 2597 times less. In addition, even with bilateral solver post-processing, PEEKABOO maintains its superior performance, demonstrating its ability to produce high-quality object masks without heavy reliance on post-processing techniques for enhanced results.

\begin{table}[!htb]
\setlength{\tabcolsep}{4pt}
\begin{center}
\caption{\textbf{Performance comparison of our method and state-of-the-art methods in unsupervised saliency detection}. ``+BS'' and ``+CRF'' indicate the utilization of the post-processing techniques~\cite{barron2016fast} and~\cite{krahenbuhl2011efficient}, respectively. The symbol $\dagger$ denotes results reproduced using the publicly available codes.}
\smallskip
\label{table:sota_saliency_detection}
\resizebox{1\textwidth}{!}{%
\begin{tabular}{lclllllllll}
\toprule[1pt]
\multicolumn{2}{c}{}
&
\multicolumn{3}{c}{DUT-OMRON}
&
\multicolumn{3}{c}{DUTS-TE}
&
\multicolumn{3}{c}{ECSSD} \\
\cmidrule(r){3-5}\cmidrule(l){6-8}\cmidrule(l){9-11}
\noalign{\smallskip}
Method & Learning & Acc & IoU & max $F_{\beta}$ & Acc & IoU & max $F_{\beta}$ & Acc & IoU & max $F_{\beta}$\\
\noalign{\smallskip}
\midrule[1pt]
\noalign{\smallskip}
HS~\cite{yan2013hierarchical} & & 84.3 & 43.3 & 56.1 & 82.6 & 36.9 & 50.4 & 84.7 & 50.8 & 67.3\\
wCtr~\cite{zhu2014saliency} & &  83.8 & 41.6 &  54.1 & 83.5 & 39.2 & 52.2 & 86.2 & 51.7 & 68.4\\
WSC~\cite{li2015weighted} & & 86.5 & 38.7 & 52.3 & 86.2 & 38.4 & 52.8 & 85.2 & 49.8 & 68.3\\
DeepUSPS~\cite{nguyen2019deepusps} & & 77.9 & 30.5 & 41.4 & 77.3 & 30.5 & 42.5 & 79.5 & 44.0 & 58.4\\
BigBiGAN~\cite{voynov2021object} & &  85.6 & 45.3 & 54.9 & 87.8 & 49.8 & 60.8 & 89.9 & 67.2 & 78.2\\
E-BigBiGAN\cite{voynov2021object} & & 86.0 & 46.4 & 56.3 & 88.2 & 51.1 & 62.4 & 90.6 & 68.4 & 79.7\\
Melas-Kyriazi et al.~\cite{melas2021finding} & & 88.3 & 50.9 & - & 89.3 & 52.8 & - & 91.5 & 71.3 & -\\
LOST~\cite{simeoni2021localizing} & &  79.7 & 41.0 & 47.3 & 87.1 & 51.8 & 61.1 & 89.5 & 65.4 & 75.8\\
DSM~\cite{melas2022deep} & & 80.8 & 42.8 & 55.3 & 84.1 & 47.1 & 62.1 & 86.4 & 64.5 & 78.5 \\
TokenCut~\cite{wang2022self} & & 88.0 & 53.3 & 60.0 & 90.3 & 57.6 & 67.2 & 91.8 & 71.2 & 80.3\\
SelfMask~\cite{shin2022unsupervised} &  \checkmark & 90.1 & \textbf{58.2} & - & 92.3 & 62.6 & - & 94.4 & 78.1 & -\\
FOUND$\dagger$~\cite{simeoni2023unsupervisedd} & \checkmark & \underline{90.7} & 57.1 & \underline{79.9} & \underline{93.5} & \underline{63.7} & \underline{85.2} & \textbf{94.9} & \textbf{80.6} & \underline{95.1} \\
DeepCut~\cite{aflalo2023deepcut} & \checkmark &  - & - & - & - & 59.5 & - & - & 74.6 & - \\
WSCUOD~\cite{lv2023weakly} & \checkmark & 89.7 & 53.6 & 64.4 & 91.7 & 59.9 & 73.1 & 92.2 & 72.7 & 85.4\\
\textbf{PEEKABOO (Ours)} & \checkmark &  \textbf{91.5} & \underline{57.5} & \textbf{80.4} & \textbf{93.9} & \textbf{64.3} & \textbf{86.0} & \underline{94.6} & \underline{79.8} & \textbf{95.3}\\ % multi
\midrule[1pt]
LOST + BS~\cite{simeoni2021localizing} & \checkmark & 81.8 & 48.9 & 57.8 & 88.7 & 57.2 & 69.7 & 91.6 & 72.3 & 83.7\\
DSM + CRF~\cite{melas2022deep} & \checkmark & 87.1 & 56.7 & 64.4 & 83.8 & 51.4 & 56.7 & 89.1 & 73.3 & 80.5\\
WSCUOD + BS~\cite{lv2023weakly} & \checkmark & 90.9 & 58.5 & 68.3 & 92.5 & 63.0 & \underline{76.4} & 92.8 & 74.2 & 89.6\\
TokenCut + BS~\cite{wang2022self} & \checkmark & 89.7 & \underline{61.8} & \underline{69.7} & 91.4 & 62.4 & 75.5 & 93.4 & 77.2 & 87.4\\
SelfMask + BS~\cite{shin2022unsupervised} & \checkmark & \underline{91.9} & \textbf{65.5} & - & 93.3 & 66.0 & - & \textbf{95.5} & \textbf{81.8} & - \\
FOUND + BS$\dagger$~\cite{simeoni2023unsupervisedd} & \checkmark & 91.7 & 60.9 & 69.1 & \underline{94.0} & \underline{66.1} & 75.0 & \underline{95.2} & \underline{81.7} & \underline{93.0} \\
\textbf{PEEKABOO + BS (Ours)} & \checkmark & \textbf{92.4} & 61.2 & \textbf{71.4} & \textbf{94.4} & \textbf{66.3} & \textbf{77.4} & 94.9 & 80.6 & \textbf{93.7} \\
\bottomrule[1pt]
\end{tabular}
}
\end{center}
\end{table}

\begin{figure*}[!htp]
\centering
\includegraphics[scale=.168]{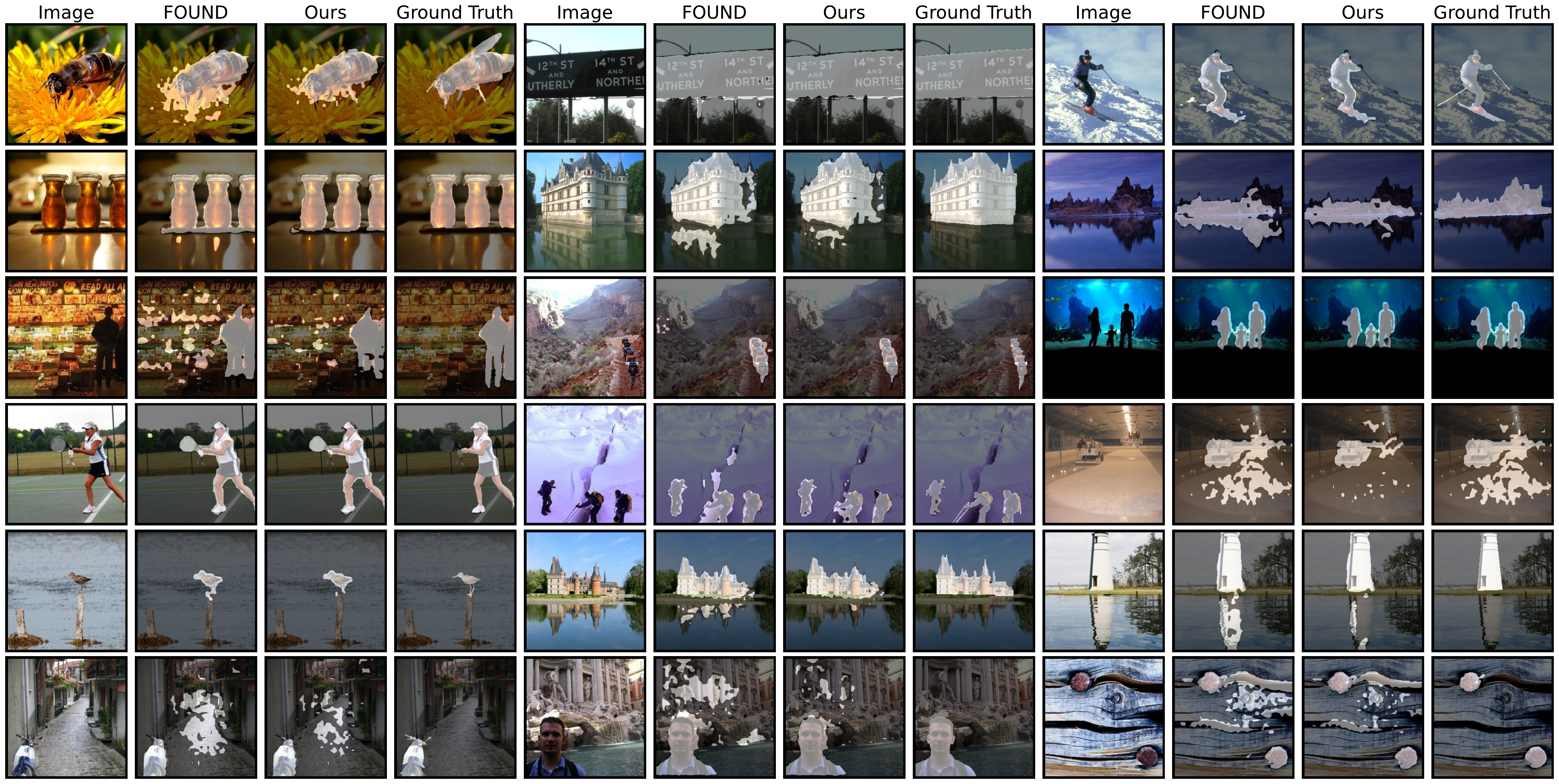}
\caption{\textbf{Visual comparison of PEEKABOO and state-of-the-art FOUND~\cite{simeoni2023unsupervisedd} on ECSSD, DUT-OMRON and DUTS-TE datasets}. Across all datasets, PEEKABOO excels in localizing salient objects, particularly when they are small, reflective, or situated against complex or dimly illuminated backgrounds. Zoom in to observe the results more closely.}
\label{Fig:qual1}
\end{figure*}

\smallskip\noindent\textbf{Qualitative Results.}\quad In Figure~\ref{Fig:qual1}, we present visual examples showcasing the predictions generated by our PEEKABOO model and the FOUND~\cite{simeoni2023unsupervisedd} baseline. The first two rows, featuring samples from the ECSSD dataset, illustrate that the baseline tends to over-segment salient objects and struggles with cases involving reflections on water and shiny surfaces, resulting in nearly duplicated segmentation. Moving to the middle two rows, which depict samples from the DUT-OMRON dataset, we observe the baseline's challenges with complex backgrounds, dark scenes, and small objects, along with its tendency to segment non-salient regions and produce noisy predictions. Finally, the last two rows, representing instances from the DUTS-TE dataset, highlight the baseline's difficulty in localizing small objects and its tendency to generate noisy and discontinuous predictions. By comparison, PEEKABOO excels in localizing salient objects under various conditions, including when they are small, reflecting on water or shiny surfaces, or in complex or low-illumination backgrounds. Moreover, PEEKABOO avoids common pitfalls such as over-segmentation of salient objects, segmentation of non-salient regions, and generation of noisy predictions. This is because MFP operates on partial inputs through masking, thereby generating context-based pixel-level representations. PCL, on the other hand, complements MFP by learning shape-level representations through aligning predictions from both unmasked and masked inputs.

\subsection{Ablation Study}
We analyze how each of the key components of the proposed framework affects the overall performance. We also examine the impact of masking on model performance.

\begin{figure}[!htb]
\centering
\setlength{\tabcolsep}{5pt}
\begin{tabular}{cc}
\includegraphics[width=2.43in]{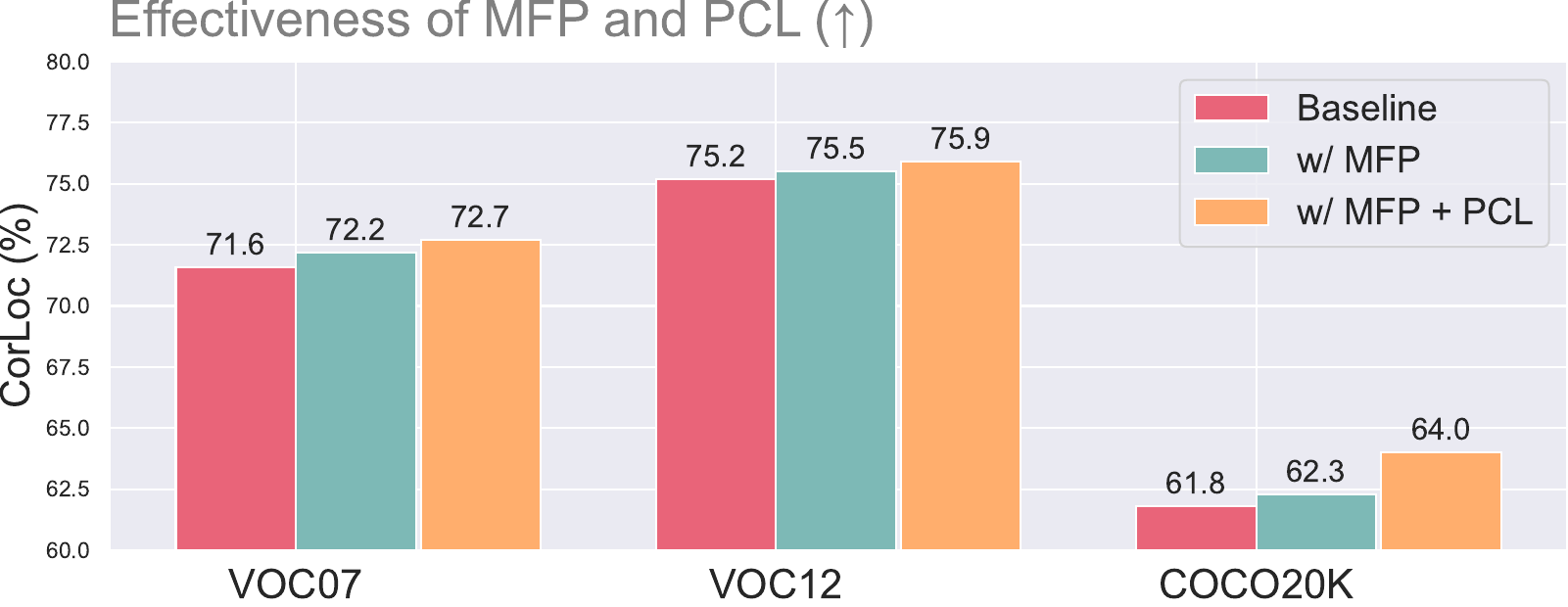} & \includegraphics[width=2.43in]{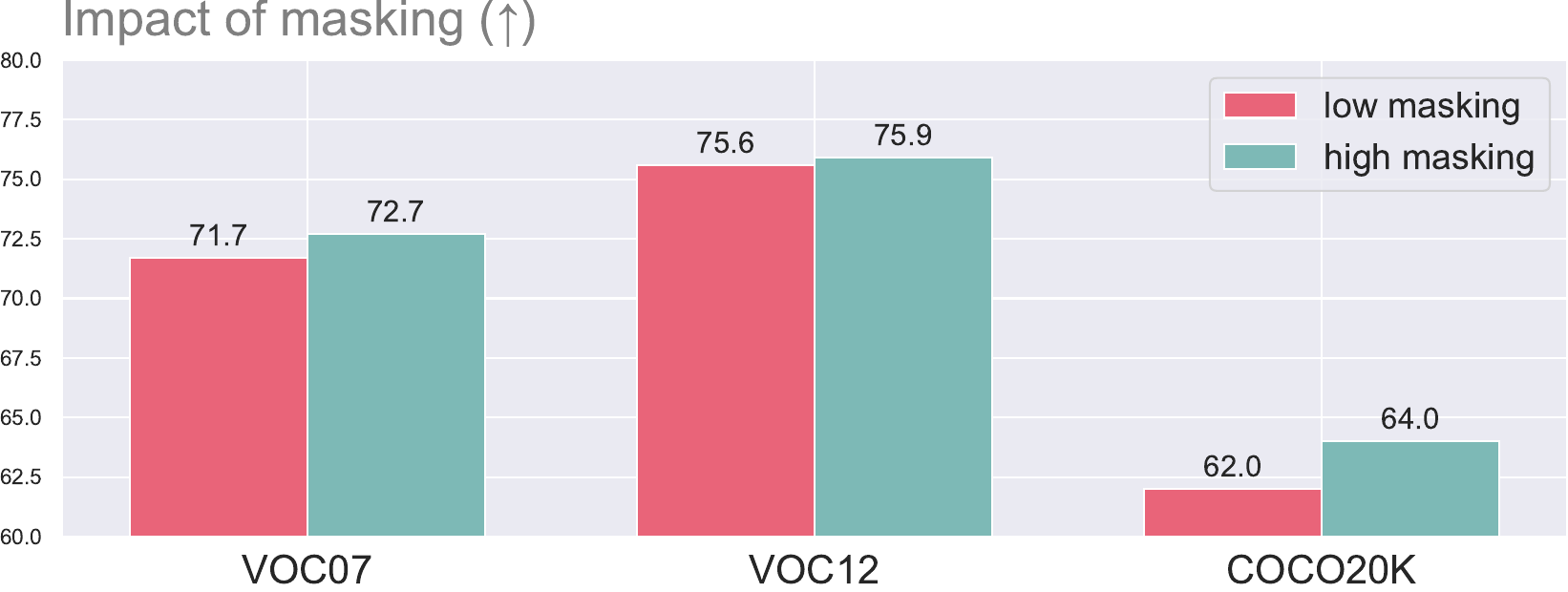}
\end{tabular}
\caption{Ablation study of different modules of PEEKABOO (left) and impact of masking (right) using VOC07, VOC12 and COCO20K datasets. MFP and PCL consistently help improve performance. PEEKABOO with high masked pixels yields better performance.}
\label{Fig:ablation}
\end{figure}

\smallskip\noindent\textbf{Effectiveness of Masked Feature Predictor (MFP).}\quad Figure~\ref{Fig:ablation} (left) illustrates the benefit of using MFP, which helps capture local semantics by generating pixel-level predictions for masked portions of the image. This enables the model to learn context by leveraging information from nearby non-masked pixels to make a prediction for the masked pixel. The effectiveness of MFP is evident in the improved localization performance across all datasets. Particularly noteworthy is the significant performance improvement observed on COCO20K, which contains a diverse range of images, including those with multiple objects, compared to the VOC07 and VOC12 datasets, which primarily consist of single large objects.

\smallskip\noindent\textbf{Effectiveness of Predictor Consistency Loss (PCL).}\quad As depicted in Figure~\ref{Fig:ablation} (left), PCL contributes to performance improvement across all datasets. By aligning the predictions of the input and its masked counterpart, we can learn representations that are invariant to masking. This enables the model to capture the shapes of objects to be segmented, resulting in more accurate predictions. Notably, the most substantial improvement is observed on COCO20K, underscoring the potential of PEEKABOO in scenarios featuring diverse images with multiple objects.

\smallskip\noindent\textbf{Impact of Masking.}\quad In Figure~\ref{Fig:ablation} (right), we illustrate the impact of masking on model performance during the training of PEEKABOO. Our training strategy involves utilizing subsets of high and low masks, where $\bm{M}_\text{scribble}$ in the subset either comprises large or small portions of zero pixels. We observe that localization performance generally improves when a large portion of the image is masked. This observation aligns with intuition since low masking preserves most of the information, leading the model to learn redundant features.

\section{Discussion and Limitations}
\noindent\textbf{Relation to Self-Supervised Learning via Dual Networks.} \quad Self-supervised learning (SSL) serves as a pre-training strategy aimed at mitigating the reliance on labeled data for supervised learning. Typically, SSL involves a pretext task that exploits the inherent semantics and structure of unlabeled data to learn the relationship between input and pseudo-label (i.e., image representations), easily generated from the input itself. Numerous SSL methods~\cite{chen2020simple,zbontar2021barlow,bardes2021vicreg,grill2020bootstrap,caron2020unsupervised,chen2021exploring} rely on Siamese networks~\cite{bromley1993signature} to encourage the similarity between two outputs, showcasing the effectiveness of self-supervised pre-training in various target tasks, including image classification, segmentation, and detection~\cite{bardes2021vicreg}.

While this similarity concept resembles the predictor consistency loss introduced in PEEKABOO, there is a key distinction in the type of output: SSL methods yield latent representations (i.e., feature vectors), whereas PEEKABOO outputs prediction labels (i.e., segmentation maps). Moreover, PEEKABOO differs from SSL methods in a number of aspects. First, SSL methods typically involve a pre-training stage on large amounts of unlabeled data followed by fine-tuning for the target task.  In contrast, PEEKABOO directly applies to the target task of unsupervised object localization. Second, while SSL methods operate on two randomly augmented samples of the image through rotation, cropping, translation, or blurring~\cite{chen2020simple,zbontar2021barlow,bardes2021vicreg}, PEEKABOO works with the original image and its masked version. Third, unlike methods such as BYOL~\cite{grill2020bootstrap}, SwAV~\cite{caron2020unsupervised} and
SimSiam~\cite{chen2021exploring}, which employ two independent networks operating on the two samples, PEEKABOO utilizes a Siamese network with shared weights, where both networks are identical. Fourth, PEEKABOO employs a completely different objective function compared to existing methods: SimCLR~\cite{chen2020simple} uses a contrastive loss requiring both positive and negative pairs, Barlow Twins~\cite{zbontar2021barlow} and VICReg~\cite{bardes2021vicreg} use a cross-correlation matrix based loss, while
BYOL~\cite{grill2020bootstrap}, SwAV~\cite{caron2020unsupervised} and SimSiam~\cite{chen2021exploring} use a cosine similarity loss. In summary, PEEKABOO aims to unify the concepts of self-supervised pretraining~\cite{chen2020simple,zbontar2021barlow,bardes2021vicreg,chen2021exploring} on unlabeled data and fine-tuning on a target task into a single framework that models contextual relationship among pixels and learns semantics and structure through image masking. This approach potentially avoids the discrepancies involved in separate pre-training and fine-tuning stages.

%---------------------------------
\newlength{\oldintextsep}
\setlength{\oldintextsep}{\intextsep}
\setlength\intextsep{0pt}
\begin{wrapfigure}{r}{0.5\textwidth}
\centering
\includegraphics[scale=0.5]{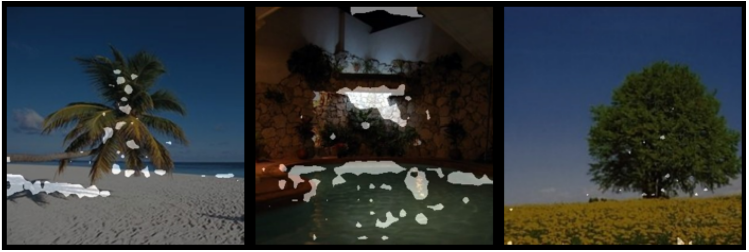}
\caption{\textbf{Visualization of failure cases of PEEKABOO on DUT-OMRON for object localization}. No refinement step is applied. Zoom in to observe the results more closely.}
\label{Fig:failure}
\end{wrapfigure}

\medskip\noindent\textbf{Limitations.} \quad Figure~\ref{Fig:failure} illustrates some instances where PEEKABOO encounters challenges. One notable limitation is observed in indoor scenes, where PEEKABOO struggles to accurately segment salient objects. Moreover, in cases where salient objects, such as trees, are easily distinguishable visually, PEEKABOO may exhibit difficulties in segmentation, suggesting potential over-generalization issues.

\section{Conclusion}
We introduced PEEKABOO, a new single-stage learning paradigm that models contextual relationship among pixels through image masking for unsupervised object localization. The proposed framework learns context-based representations at the pixel-level by making predictions on masked images and at the shape-level by matching the predictions of the masked input to the unmasked one. Through extensive experiments, our quantitative and qualitative results demonstrated that our approach outperforms state-of-the-art methods, excelling in localizing salient objects, especially when they are small, reflect on water or shiny surfaces, or when the background is complex or poorly illuminated. Notably, our method avoids over-segmenting salient objects, segmenting non-salient regions, and producing noisy predictions. For future work, we aim to explore learning context-based representations for direct localization of individual concepts in open-vocabulary semantic segmentation.

\bigskip\noindent\textbf{Acknowledgments.}\quad This work was supported in part by the Discovery Grants Program of the Natural Sciences and Engineering Research Council of Canada under grant RGPIN-2024-04291.
\bibliography{references}

%==========================================================================================
\clearpage
\setcounter{page}{1}
\section*{----- Supplementary Material -----}

\section{Implementation Details}
\noindent\textbf{Data preprocessing.}\quad Images and masks are resized to $224 \times 224$ and normalized using mean and standard deviation of ImageNet. Similar to SelfMask~\cite{shin2022unsupervised}, we apply basic data augmentation techniques, including random scaling within the range of $[0.1, 3.0]$ and Gaussian blurring with a probability of 0.5. The parameters of the bilateral solver are the same as those provided in~\cite{barron2016fast}.

\smallskip\noindent\textbf{Architecture.}\quad We construct PEEKABOO by using a frozen ViT-S/8~\cite{dosovitskiy2020vit} architecture pretrained using DINO~\cite{caron2021emerging} as an encoder for feature extraction, from the last attention layer, with a lightweight segmenter head that is a single $1\times 1$ convolutional layer
decoder having only 770 learnable parameters.

\begin{figure}[!htp]
\centering
\includegraphics[scale=.33]{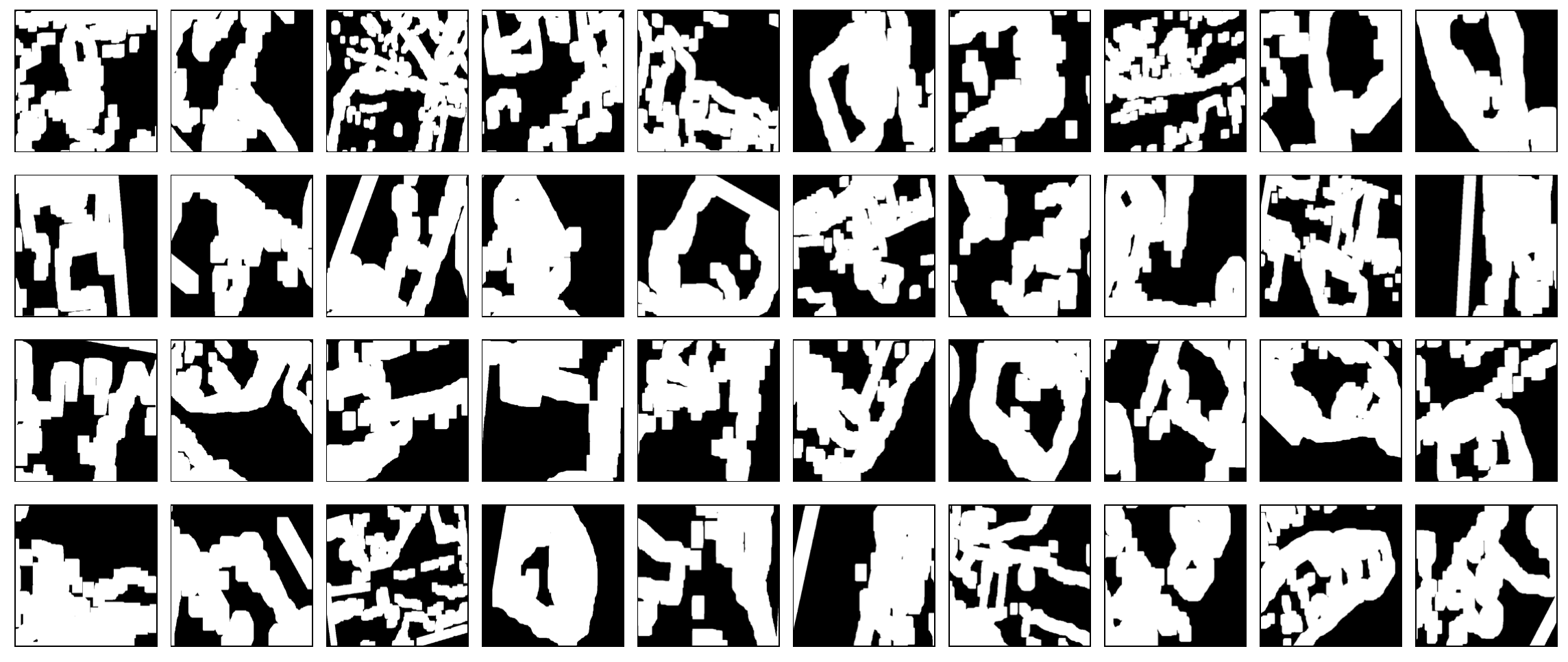}
\caption{\textbf{Visualization of masks during training in PEEKABOO}. Some masks cover more than 50\% of the image. Images are from Irregular Masks Dataset~\cite{liu2018image} after applying binary thresholding.}
\label{Fig:mask}
\end{figure}

\smallskip\noindent\textbf{Model Training.}\quad PEEKABOO is trained in a single stage, requiring only a collection of images. We use the Adam optimizer with
a learning rate schedule to minimize the total loss function. In addition to the total loss, we also compute the binary cross-entropy loss between the raw predicted masks from the unsupervised segmenter branch and their binarized version. This encourages the predicted soft masks to closely resemble their binarized counterparts. We set the trade-off hyperparameter to 4. The model is trained for 500 iterations on only $10,553$ images from the DUTS-TR dataset~\cite{wang2017learning} with a batch size of 50, which corresponds to slightly more than 2 epochs. For image masking, we utilize the Irregular Mask (IMs) dataset~\cite{liu2018image}, containing masks with random streaks and holes of various shapes, as illustrated in Figure~\ref{Fig:mask}.

\smallskip\noindent\textbf{Model Testing.}\quad After training, given an input image, the model simply makes a prediction by outputting a segmentation mask for the salient object(s). During inference, the input image undergoes normalization only. Moreover, there is no random masking procedure applied, as is done during the training stage.

\smallskip\noindent\textbf{Hardware and software details.}\quad The experiments were performed on a Linux workstation running 4.8Hz and 64GB RAM, equipped with a single NVIDIA RTX 3080Ti GPU featuring 12GB of memory. All algorithms are implemented using the PyTorch framework.

\section{Additional Visual Comparison Results}
\begin{figure}[!htp]
\centering
\includegraphics[scale=.167]{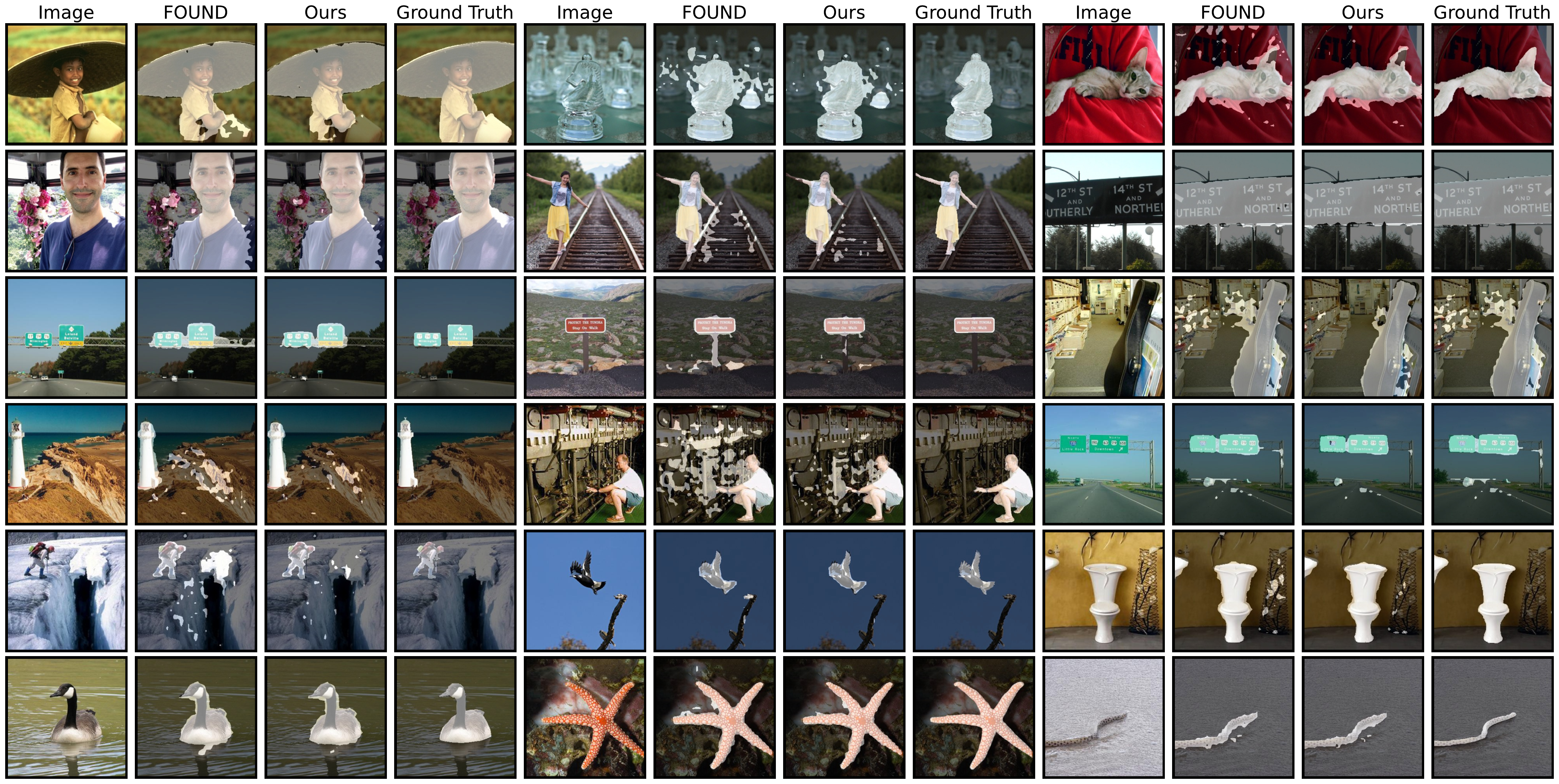}
\caption{\textbf{More examples of visual comparison of PEEKABOO and state-of-the-art FOUND~\cite{simeoni2023unsupervisedd} on ECSSD, DUT-OMRON and DUTS-TE datasets}. Across all datasets, PEEKABOO excels in localizing salient objects, particularly when they are small, reflective, or situated against complex or dimly illuminated backgrounds. Zoom in to observe the results more closely.}
\label{Fig:qual_ext}
\end{figure}

In Figure~\ref{Fig:qual_ext}, we present additional experimental results on unsupervised object localization to further demonstrate the effectiveness of PEEKABOO in localizing salient objects. As can be see, PEEKABOO consistently excels in localizing salient objects across all datasets. Its performance is particularly noteworthy when dealing with small objects, reflective surfaces, and objects situated against complex or dimly illuminated backgrounds. This capability highlights PEEKABOO's robustness and adaptability in various challenging scenarios, further validating its superiority over strong baselines in unsupervised object localization tasks.

\section{Method Cost Discussion}
We compare PEEKABOO against training-free and training-based methods, which we find have significantly different costs at both training and inference time. Specifically, we demonstrate the efficiency of our method. PEEKABOO is a segmenter head, on top of a frozen Self-distillation with No Labels (DINO)~\cite{caron2021emerging} as an encoder, which consists of a lightweight \textbf{single $1\times 1$ convolutional layer} decoder having only \textbf{770 learnable parameters}. The model is trained for \textbf{2 epochs} on \textbf{$10,553$ images} from the DUTS-TR dataset~\cite{wang2017learning} on a \textbf{single consumer grade NVIDIA RTX 3080Ti GPU}.

Inference with training-free methods like TokenCut~\cite{wang2022self} and Deep Spectral Methods (DSMs)\cite{melas2022deep} is slowed down due to the expensive computation of the Laplacian matrix eigenvectors. LOST\cite{simeoni2021localizing}, while somewhat faster, notably lags behind PEEKABOO in terms of localization performance.

Among training-based methods, FreeSOLO~\cite{wang2022freesolo} stands out with approximately 66 million learnable parameters, trained over 241 thousand unlabeled images for 60 thousand iterations across 8 GPUs, making its training considerably more resource-intensive compared to ours. Also, its backbone is pretrained on ImageNet with 1.28 million unlabeled images. SelfMask~\cite{shin2022unsupervised} utilizes 36 million learnable parameters and trains for 12 epochs on the DUTS-TR dataset~\cite{wang2017learning}. COMUS~\cite{zadaianchuk2022unsupervised} requires three days of training on two 8-GPU servers for its heavy segmentation backbone. DINOSAUR~\cite{seitzer2022bridging} is trained on over 300 thousand images from synthetic and real-world sources and demands 8 GPUs for training. DeepCut~\cite{aflalo2023deepcut} has 30K learnable parameters, and WSCUOD~\cite{lv2023weakly}, which incorporates a DINO-ViT-S/16 backbone, consists of 2 million learnable parameters and requires training on 6 GPUs, making it substantially more expensive to train compared to our method.

\end{document}